\documentclass[letterpaper, 10 pt, conference]{ieeeconf}
\usepackage{epsfig,epsf,rotating,setspace,latexsym,amsmath,amssymb,amsfonts,bm,theorem,subfigure,epstopdf,cite,authblk,bbm,nonfloat,comment}
\usepackage{algorithm, algorithmic}
\usepackage{color}
\usepackage{mathtools}
\usepackage{soul}
\usepackage[most]{tcolorbox}
\usepackage{xcolor}

\IEEEoverridecommandlockouts
\allowdisplaybreaks
\begin{document}

\title{RunAgent: Interpreting Natural-Language Plans with Constraint-Guided Execution} 

\author{
    Arunabh Srivastava$^{1}$, Mohammad A. (Amir) Khojastepour$^{2}$, Srimat Chakradhar$^{2}$, Sennur Ulukus}
\affil[1]{\normalsize University of Maryland, College Park, MD}
\affil[2]{\normalsize NEC Laboratories America, Inc.}

\maketitle

\begin{abstract}
     Humans solve problems by executing targeted plans, yet large language models (LLMs) remain unreliable for structured workflow execution. We propose RunAgent, a multi-agent plan execution platform that interprets natural-language plans while enforcing stepwise execution through constraints and rubrics. RunAgent bridges the expressiveness of natural language with the determinism of programming via an agentic language with explicit control constructs (e.g., \texttt{IF}, \texttt{GOTO}, \texttt{FORALL}). Beyond verifying syntactic and semantic verification of the step output, which is performed based on the specific instruction of each step, RunAgent autonomously derives and validates constraints based on the description of the task and its instance at each step. RunAgent also dynamically selects among LLM-based reasoning, tool usage, and code generation and execution (e.g., in Python), and incorporates error correction mechanisms to ensure correctness. Finally, RunAgent filters the context history by retaining only relevant information during the execution of each step. Evaluations on Natural-plan and SciBench Datasets demonstrate that RunAgent outperforms baseline LLMs and state-of-the-art PlanGEN methods.
\end{abstract}

\section{Introduction}\label{sec: introduction}

Problem solving or task completion fundamentally rely on devising and executing a targeted plan. Significant advancements in artificial intelligence (AI) and particularly in Large Language Models (LLMs) have enabled machines to emulate sophisticated human cognitive processes, such as identification and reasoning, that are collectively referred to as human cognitive processes. With the advent of LLMs, the development of AI agents to perform specialized tasks has become one of the primary research frontiers. However, as LLMs transition from conversational chatbots to autonomous multi-agent systems which execute complex multi-step tasks, the need for a structured framework becomes critical to navigate these challenges. We define the planning and execution of workflows as a structured process to answer questions, solve problems or facilitate the design of AI agents. To this end, we introduce the execution platform \emph{RunAgent} which extends the body of works on planning, generation, and execution of workflows \cite{parmar2025plangen,rawat2025preactmultistepplanningreasoning, wei2025plangenllms, kaelbling1998planning}. RunAgent performs the step-by-step execution of a workflow, rigorously checks the output of each step for validity with respect to rubrics and constraints, and verifies the output. 

The RunAgent platform comprises key architectural features. First, RunAgent allows for human-in-the-loop (HITL) operations: (a) User-specification: This feature allows the user to furnish desired workflows, constraints, facts or specifications, and rubrics; and (b) User-feedback: The HITL feature also allows the user to provide feedback on an operation, enabling users to audit the log and provide feedback on steps of the workflow, the final output or other particular parts of the log. Second, RunAgent performs autonomous generation of relevant constraints and verifies each step of the workflow. This means that RunAgent does not rely solely on user-specification or user-feedback to recognize constraints. Third, RunAgent employs a flexible language that seamlessly reconciles the formal determinism of programming languages and the expressivity of natural language. The language for AI agent development should be capable of performing adaptive reasoning, which means that it can significantly benefit from using the expressive freedom of natural language. This implies that rigid code cannot handle context-specific exceptions, diminishing the adaptive intelligence required for effective AI agent operations. For example, when context-specific circumstances arise, new exceptions are given, or priorities shift, a rigid-language cannot handle these cases. Conversely, using the expressivity of natural language in defining a workflow also has limitations. A workflow may not always be a series of sequential steps and may have non-linear control flow. Hence, while the flexibility of interpreting natural plan directives must be maintained, there is a need for the ability to execute a series of steps or perform branch operations in a principled and deterministic manner with formal execution guarantees. 

The remainder of the paper is organized as follows. Related works are presented in Section~\ref{sec:related-work}. We then describe our agentic language and its core constructs in Section~\ref{sec:language}. The design and architecture of the RunAgent platform is discussed in Section~\ref{sec:RunAgent}. In Section~\ref{sec:evaluation}, we evaluate RunAgent against baseline LLMs and state-of-the-art algorithms. Finally, we provide supplementary details and additional experimental results in the Appendix.

\section{Related Work}\label{sec:related-work}
To support reliable plan execution, several tool-augmented LLM agent frameworks have been proposed. Systems such as AutoGen \cite{wu2024autogen}, Voyager \cite{wang2023voyager}, and LLM-generated plan heuristics \cite{correa2025classical} demonstrate that delegating sub-tasks to programmatic or symbolic executors improves the accuracy of multi-step plan execution. Building on this paradigm, RunAgent introduces LLM-driven control flow constructs, including loops, conditionals, action modules, and Python execution, to robustly execute LLM-generated plans. More recent work, such as Magentic UI \cite{mozannar2025magentic}, explores plan generation with HITL interaction via multi-agent orchestration, co-planning, co-execution, action guards, and memory, but relies heavily on human feedback for verification. XPF \cite{Rao25XPF} presents an agentic AI system for business workflow automation that integrates planning, execution, and verification with HITL plan editing. In contrast, RunAgent performs constraint generation and constraint checking based on grounded facts, in addition to performing syntactic and semantic verification at each step. RunAgent further determines the most appropriate execution strategy for each step, selecting between tool usage, code generation and execution, or direct LLM reasoning. Moreover, RunAgent is capable of reasoning and generating code across single or multiple steps or sub-steps. Another feature of RunAgent is the filtering of context history prior to providing it to the LLM for step execution.

\section{Agentic Language}\label{sec:language}
AI agents are designed to augment or automate expert functions, enabling domain specialists to develop the AI agents themselves without traditional programming expertise. Ideally, defining an agent should be as simple as writing a natural-language specification, referred to as a \emph{task}, without extensive algorithmic thinking. During execution, the agent receives data and natural-language directives, collectively forming a task \emph{instance}. These directives are interpreted by the agent and may alter its operation by introducing new circumstances, priorities, or exceptions, distinguishing them from data or signals processed by the agent. Together, the task and instance govern how the workflow implementer translates expert intent into reliable execution.

However, while this natural language description provides inherent adaptability, it lacks the formal guarantees required for robust and reliable automation. Specifically, unrestricted workflows fail to capture structured control flow such as branching and iteration. RunAgent addresses these limitations by employing a flexible agentic language that bridges the rigidity of programming languages and the expressivity of natural language. Through a small set of reserved keywords, RunAgent preserves natural-language flexibility while enabling deterministic execution. Next, we introduce these keywords and motivate their design.

To ensure robust execution, a workflow must be decomposed into a series of ordered steps that account for both linear and conditional logic. While simple workflows follow a single path, more complex workflows involve judgments that induce branching or loops, yielding a directed graph structure. In this model, each node executes an action, and outgoing edges are selected based on the step outcome, motivating the introduction of conditional (\texttt{IF}) and branching (\texttt{GOTO}) instructions in the flowgraph.

In addition, our observation shows that LLMs struggle with the \emph{for all} operation (e.g., do this task \emph{for all} items in a set). In most cases, the LLM only performs the operation on a partial subset instead of all items in the set. Hence, we introduce the \texttt{(FORALL)} instruction to overcome this shortcoming and ensure iterative steps are implemented comprehensively. 

LLMs are generally more reliable at performing algorithmic or computational operations by generating code than by directly producing results. Accordingly, RunAgent selects between code-based and LLM-based execution for each step. In order to facilitate deterministic execution, \texttt{(PYTHON)} and \texttt{(LLM)} instructions explicitly direct RunAgent to execute a step via generated Python code or direct LLM invocation, respectively. RunAgent also determines whether an existing tool should be used to execute a step; this decision can be overridden with the \texttt{(TOOL)} instruction, which forces RunAgent to select and invoke an appropriate tool. By providing these keywords, RunAgent empowers domain experts to decide the execution modality at each step, ensuring that each step of the workflow is executed using the most reliable implementation method.

RunAgent explicitly detects keyword instructions at each step to ensure deterministic execution, rather than leaving their interpretation to the LLM. In summary, by using the introduced keyword instructions, the workflow has a hybrid structure, enforcing the algorithmic structure of programming languages while maintaining human-like and open-vocabulary descriptions for the steps.

Our vision of an agentic language differs from the formal definition of programming languages and may arguably not be perceived as a computer language, as it does not follow a rigid syntax. In our proposed language, there is a series of steps (and sub-steps) written in free-form natural language, and it is the role of an interpreter to detect keywords. We aim to minimize the use of keywords; however, there is the possibility of extending them as it becomes evident that they are essential. We recognize that such a language places the burden of detecting keywords, as well as interpreting them and potentially generating multiple sub-steps (e.g., for \texttt{(FORALL)}), on the execution platform; however, it enables flexible and simple natural language statements for each step. This differs from agentic languages such as V{\"o}l \cite{voel}, in which a strict syntax is enforced and any deviation from it results in an error.

\section{RunAgent Design}\label{sec:RunAgent}
\begin{figure*}[t]
    \centering
    \includegraphics[width=\linewidth]{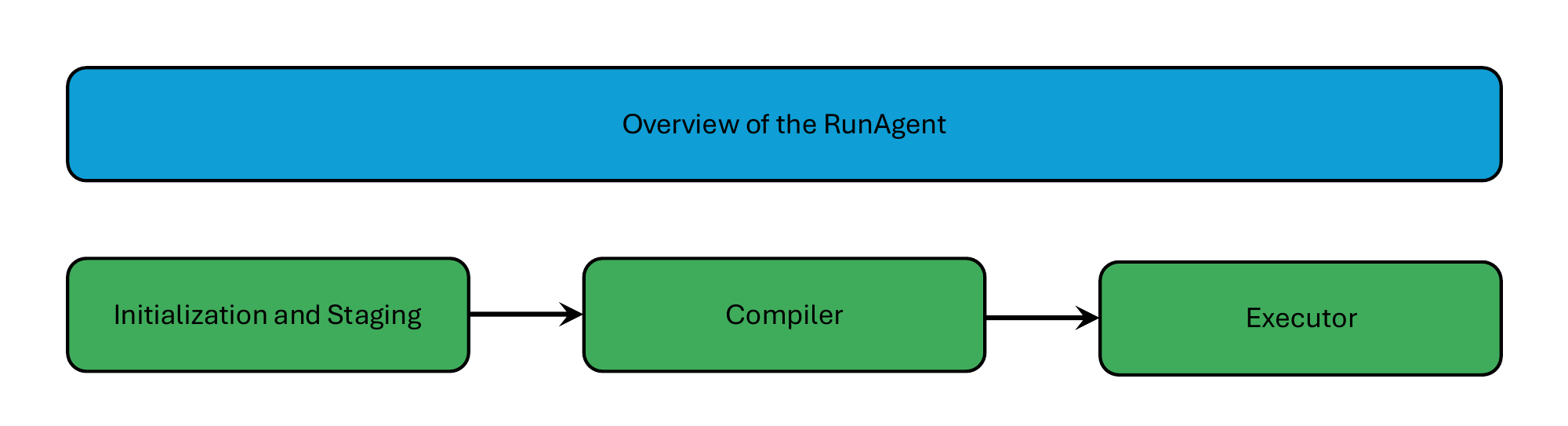}
    \caption{An overview of RunAgent, highlighting its three main modules.}
    \label{fig: overview block diagram}
\end{figure*}

\begin{figure}
    \centering
    \includegraphics[width=\linewidth]{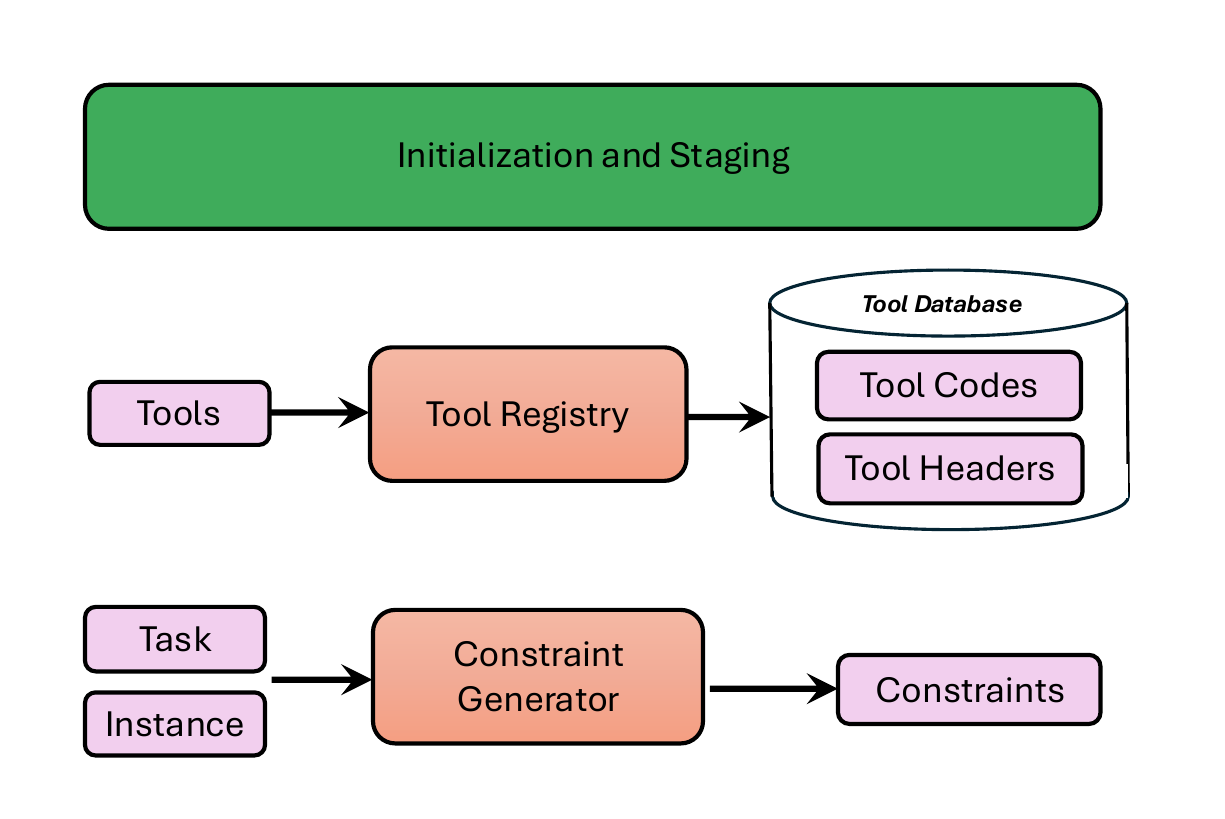}
    \caption{A description of the Initialization and Staging module. This is described in Sec.~\ref{subsection: Tool Registry Setup and Constraint Generation}.}
    \label{fig: init and staging block diagram}
\end{figure}

RunAgent is a multi-agent LLM system that takes a task, an instance of the task, and a step-by-step plan to solve the task-instance pair as input, and performs a step-by-step implementation of the plan. RunAgent employs several specialized LLM blocks to execute essential cognitive actions such as identification, judgment, and reasoning, where human-like nuance is required. Each LLM block is designed to perform a particular task that is specialized through targeted prompting and is provided with a selected subset of the conversation history and relevant contextual information. At the start of plan implementation, RunAgent employs an LLM reasoner block that uses the task–instance pair to derive constraints that must hold for the plan execution to be correct.

At a fundamental level, RunAgent implements the plan step-by-step in a sequential fashion by utilizing previously implemented steps, the task, the instance, and tool details to solve the next step. At each step, LLM judge blocks in RunAgent decide whether (i) the LLM, (ii) the Python code generator and then code executor, or (iii) a tool from the tool database should be used for execution of this step. The execution selection method may also be overridden, and the execution method can be directly specified with instructions \texttt{LLM:}, \texttt{PYTHON:} or \texttt{TOOL:} written as a prefix to the step description. Once a step is implemented, it is validated against a subset of relevant constraints, as well as rubrics. These relevant constraints are extracted from the set of constraints which are generated at the start. Additional constraints or rubrics can also be specified at the beginning or as user feedback. If verification fails, RunAgent makes more attempts to execute the step by providing the log of the previous actions and associated errors.

As discussed earlier, predefined keywords \texttt{IF}, \texttt{GOTO} and \texttt{FORALL} can also be used to initiate branches, jumps and iterative procedures at any step. In addition, RunAgent performs error correction at each step. In order to move to the next step, the output of the current step must not have any programming errors and all constraints relevant to the step must be satisfied. If the step execution fails more than a set number of times, then RunAgent falls back to an LLM response, so that the plan can be implemented completely. We also maintain detailed logs of the plan implementation, which list all the step outputs, errors and intermediate generation steps. In the following, we discuss the implementation details and different features of RunAgent.

\subsection{RunAgent Structure}

At a high level, RunAgent comprises three main blocks as shown in Fig.~\ref{fig: overview block diagram}: (i) Initialization and Staging, (ii) Compiler, and (iii) Executor blocks. 

The Initialization and Staging block depicted in Fig.~\ref{fig: init and staging block diagram} is responsible for setting of parameters, tool registration in the form of code or binary with the descriptions, and other structural methods, including extraction of embeddings that are used in retrieving the relevant tools for the execution of a task. When RunAgent is invoked to perform a \emph{task} for a specialized \emph{instance} using a \emph{plan}, the staging also comprises of setting up the facts and constraints. The facts and constraints may be given as input as a part of the instance. In addition, the set of constraints is derived by the Staging module of RunAgent from the task and instance to ensure comprehensive coverage of all constraints.

\begin{figure}
    \centering
    \includegraphics[width=\linewidth]{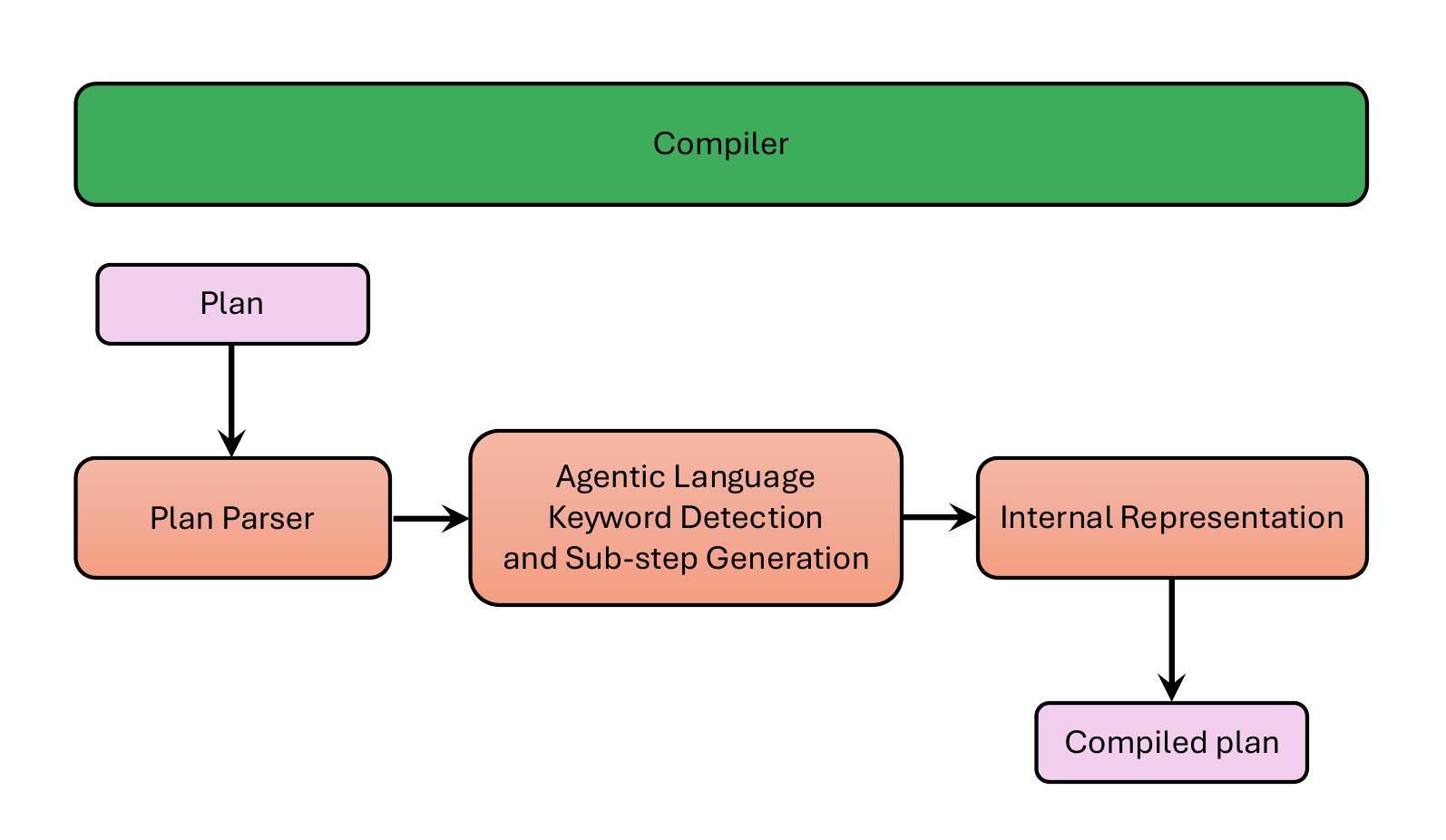}
    \caption{A description of the Compiler module. This is described in Sec.~\ref{subsection: compiler module}.}
    \label{fig: compiler block diagram}
\end{figure}

The next main block is the Compiler, depicted in Fig.~\ref{fig: compiler block diagram}. This block takes the plan as input and generates the compiled plan as output by first parsing the plan into a Python dictionary structure, identifying the agentic language keywords and appropriately generating the sub-steps, for operations like ``FORALL", and finally generating an internal representation of the steps that are used for execution by the Executor block.

\begin{figure*}
    \centering
    \includegraphics[width=0.8\linewidth]{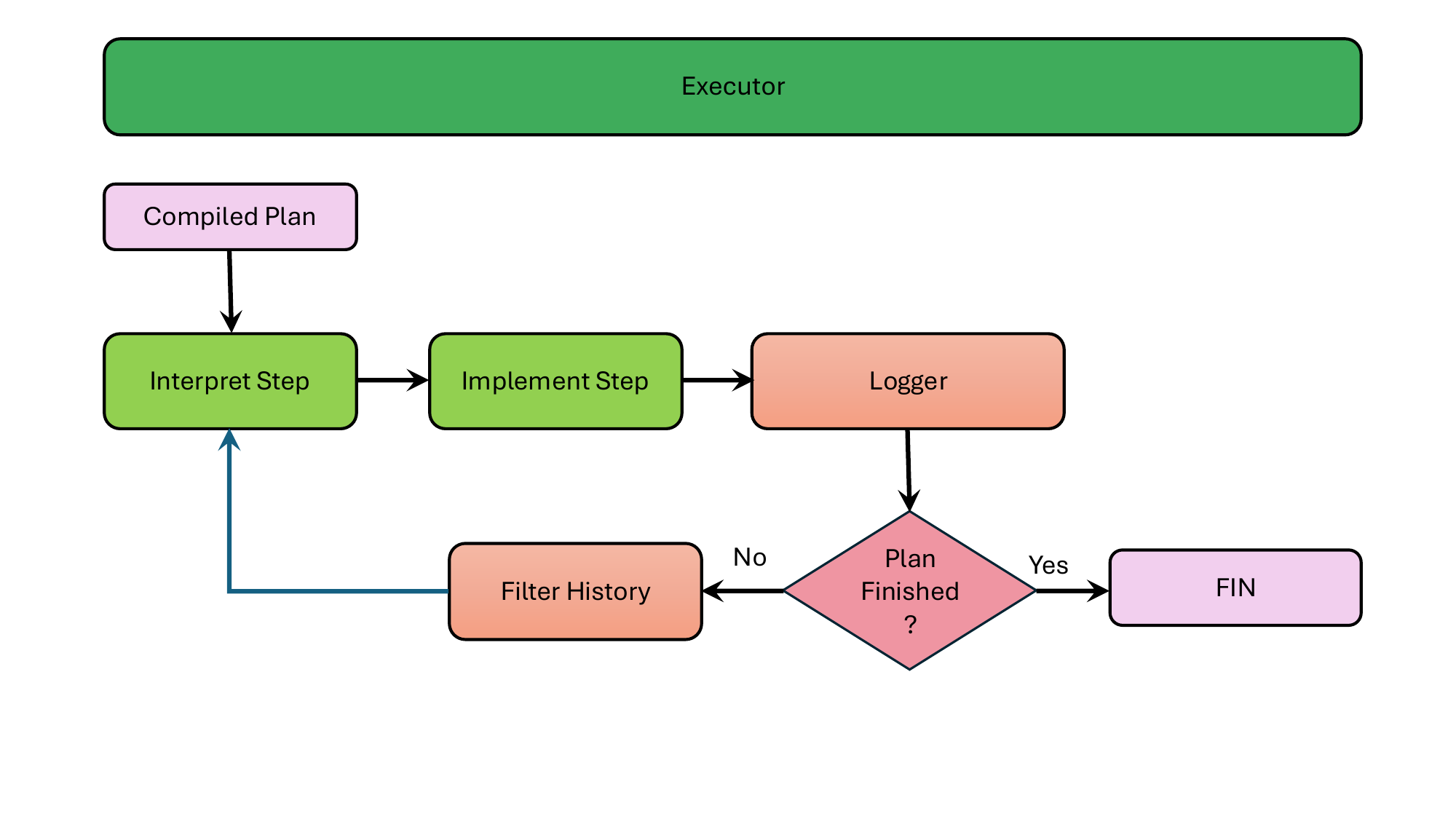}
    \caption{A description of the Executor module. It is described further in Sec.\ref{subsection: execution module}.}
    \label{fig: executor block diagram}
\end{figure*}

Finally, the Executor block depicted in Fig.~\ref{fig: executor block diagram} takes the compiled plan and performs the operation directed by each step by sequentially interpreting the step, implementing the step, logging the results of the execution of the step and filtering the relevant part of the execution context to be used in the execution of the next step, carrying on until the plan is completely executed.

\subsection{Initialization and Staging Module}\label{subsection: Tool Registry Setup and Constraint Generation}
The initialization and staging module performs two main tasks: Tool Registry Setup and Constraint Generation. All tools are stored as Python functions, which can be implemented via subroutines during execution. The tools are stored in a registry and a description is generated for each tool so that the LLM-based tool-calling module can utilize these function descriptions to determine which tool to use to execute a step. Each tool is stored as source code, e.g., Python functions, and the required description and information for its execution is also provided and is available to RunAgent. Moreover, a set of constraints is generated at this stage using the task and instance. In order to facilitate easy constraint checking, atomic constraints, i.e., constraint which represents a single, irreducible logical condition, are generated. These constraints are later filtered depending on the step, and the selected ones are verified during the step's execution to ensure correct plan execution.

\subsection{Compiler Module}\label{subsection: compiler module}
The compiler module takes the natural language plan as input, parses the plan, detects the agentic language keywords, generates corresponding sub-steps, and stores the plan in an internal structure.

\subsubsection{Plan Parsing}
Before a plan is implemented in RunAgent, it is parsed appropriately into a machine-readable format. This format is a list of Python dictionaries, where each dictionary contains the step number, the text of the step, and the reasoner hint, which could be \texttt{(LLM)}, \texttt{(PYTHON)}, \texttt{(TOOL)} or None. The parser also incorporates the inclusion of sub-steps of the form $1.1$,$2.3.1$, etc. A final step is then appended to indicate that the plan has been completed.

\subsubsection{Agentic Language Interpreter}
This module detects agentic-language keywords such as \texttt{IF}, \texttt{GOTO}, and \texttt{FORALL}, and generates the required sub-steps accordingly. It governs how steps are implemented based on these keywords, including how step outputs are routed, how the running state summary is constructed, and how the execution context is managed for intermediate processes. The resulting information is stored in an internal compiled plan structure.

In the following we describe the way in which keywords are interpreted and prepared by the Compiler module and later processed by the Executor module.
\begin{enumerate}
    \item \textbf{GOTO}: In RunAgent, the \texttt{GOTO} keyword is an unconditional jump, and it is internally stored with the current step number, the keyword \texttt{GOTO} and the step number to which the jump must be initiated. Whenever a \texttt{GOTO} keyword is encountered by the Executor module, it adds the target step to the execution context and moves to targeted step number.
    \item \textbf{IF}: Conditional steps are encoded as natural language \texttt{IF} statements. RunAgent utilizes an LLM identification block to parse the step that contains the \texttt{IF} keyword into an internal structure comprised of the current step number, the keyword \texttt{IF}, the condition, and the instruction that must be carried out if the condition is satisfied. When the Executor Module encounters \texttt{IF} step, an LLM Judge block is used to check the condition based on the execution context. If the condition is satisfied, then the instruction from the \texttt{IF} statement is appended to the execution context and sent for step implementation. Otherwise, the Executor moves to the next step. Notably, the \texttt{GOTO} keyword can be a part of the instruction in an \texttt{IF} step. This facilitates implementation of conditional jumps or conditional loops.
    \item \textbf{FORALL}: The \texttt{FORALL} keyword is used to specify that a step instruction should be executed for all items in an iterative list. Its implementation is similar to \texttt{IF}, where an LLM identifier block is used to obtain and update the internal step representation with the list of iterates or the hint that is required to pull the list of iterates based on the execution context. This makes it possible for the \texttt{FORALL} statement to refer to iterates based on previously executed steps and their outputs. The LLM identifier block is also used to strip the associated instruction from the step definition that has to be performed for each iterate and update the internal representation of this step. The Executor module performs the instruction for each iterate as a sub-step by temporarily augmenting the execution context and storing the resulting sub-step output. Once the outputs for all iterates are obtained, the individual outputs from the sub-steps are aggregated to form the output of the current \texttt{FORALL} step instruction that is then permanently appended to the execution context.
\end{enumerate}

\subsubsection{Plan Implementation Representation}\label{subsection: Plan Implementation Representation}
In order to ensure the correct implementation of each step based on the known information and outputs of the previously successfully implemented steps, all the information is logged in a special format, which we call the execution context. The task, instance, incoming step formats and implementation instructions to the LLM are given first. Then, a JSON structure consisting of the tool names, their inputs and outputs and their natural language descriptions are given as well. Then, the step instructions and outputs are added in an XML format. Each step instruction including the previous steps and the current steps is written in the format: $<$step$>$$<$number$>$number of the step$<$/number$>$$<$instruction$>$Step instruction $<$/instruction$>$$<$/step$>$. Similarly each step output is written in the following format: $<$step$>$$<$number$>$number of the step$<$/number$>$$<$output$>$Step Output $<$/output$>$$<$/step$>$.

The Executor module uses this internal representation, which comprises the step representation as described above, along with the plan state; while the constraints and the facts that are relevant to the step are extracted and cached until the step is executed successfully. In addition, a state summary of the execution output of the plan and relevant facts is also maintained in the internal representation and updated whenever a step is executed successfully.

\begin{figure*}
    \centering
    \includegraphics[width=0.8\linewidth]{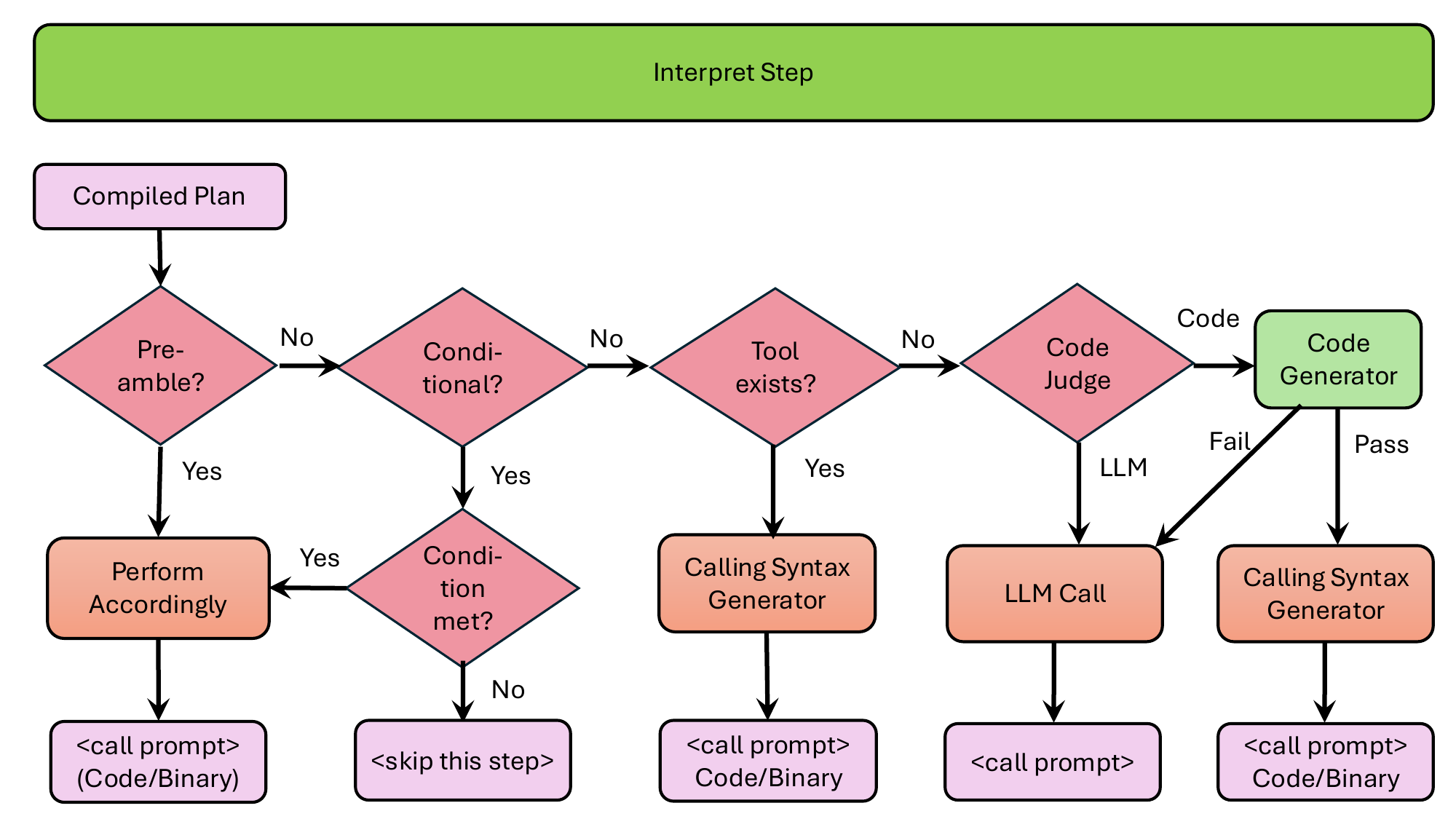}
    \caption{A description of the Interpret Step module, which is a part of the Executor module. It is described further in Sec.~\ref{subsection: Step Interpretion Methods}.}
    \label{fig: interpret step block diagram}
\end{figure*}

\subsection{Executor Module}\label{subsection: execution module}
The Executor Module takes the compiled plan as input, and interprets, implements, and logs the output of each step until the plan is finished. The Interpret Step module realizes the correct implementation of different keywords and directives, checks for preference in using tools, programming (generating and executing code), or direct LLM execution of the instructions. The Implement Step module executes the step, performs a sanity check and semantic verification of the output, and performs additional constraints and rubric checking for each step instruction. The Executor module moves to the next step once the current step is successfully performed and all the issues with the current step have been resolved or a maximum number of trials has been reached. 

During execution, the executor maintains the execution context, implemented as an ordered message history, containing (i) the system specification and all previously accepted step instruction–output pairs which are described in Section~\ref{subsection: Plan Implementation Representation}, and (ii) a result log which stores the details about each step in a readable format. In addition, the executor also maintains a running state summary which stores short summaries of the output for each step. This running summary is used for constraint validation at each step.

\subsubsection{Interpret Step Module}\label{subsection: Step Interpretion Methods}
The block diagram of Interpret Step module is shown in Fig.~\ref{fig: interpret step block diagram}. The module first checks if the step has an implementation prefix \texttt{(LLM)}, \texttt{(PYTHON)} or \texttt{(TOOL)}, or has no prefix at all. This module provides the Implement Step module with a ``call prompt", if the step needs to be implemented, or a $<$``skip this step"$>$ prompt so that the Implement Step module can skip this step. We now discuss how the call prompt is generated for different types of steps.

If the step has no prefix, then an LLM tool-judge block is given the execution context and decides if a tool can be used to implement this step. If a tool from the tool registry can be used to implement the step, then the subroutine is called locally to produce the output of the step. Otherwise, an LLM Python-judge block is invoked to determine whether the step should be executed in Python or by the LLM, based on the execution context and the nature of the task: Python is preferred for algorithmic steps, whereas the LLM is better suited for tasks requiring creativity or natural language processing. If the step has a prefix, then the Python-judge is not called, and the step is implemented strictly using the prefix method.

If the step has the \texttt{(TOOL)} prefix, or it is determined that a tool is the best option to implement the step, tool selection uses standard algorithms such as similarity matching, direct LLM calls with the tool registry database, or more advanced methods such as TacTool \cite{TacTool}. Next, the interpreter uses the execution context to generate the call prompt, and the module's output consists of the call prompt, which specifies how the tool should be invoked, along with the corresponding tool in the form of code or a binary.

\begin{figure*}
    \centering
    \includegraphics[width=0.8\linewidth]{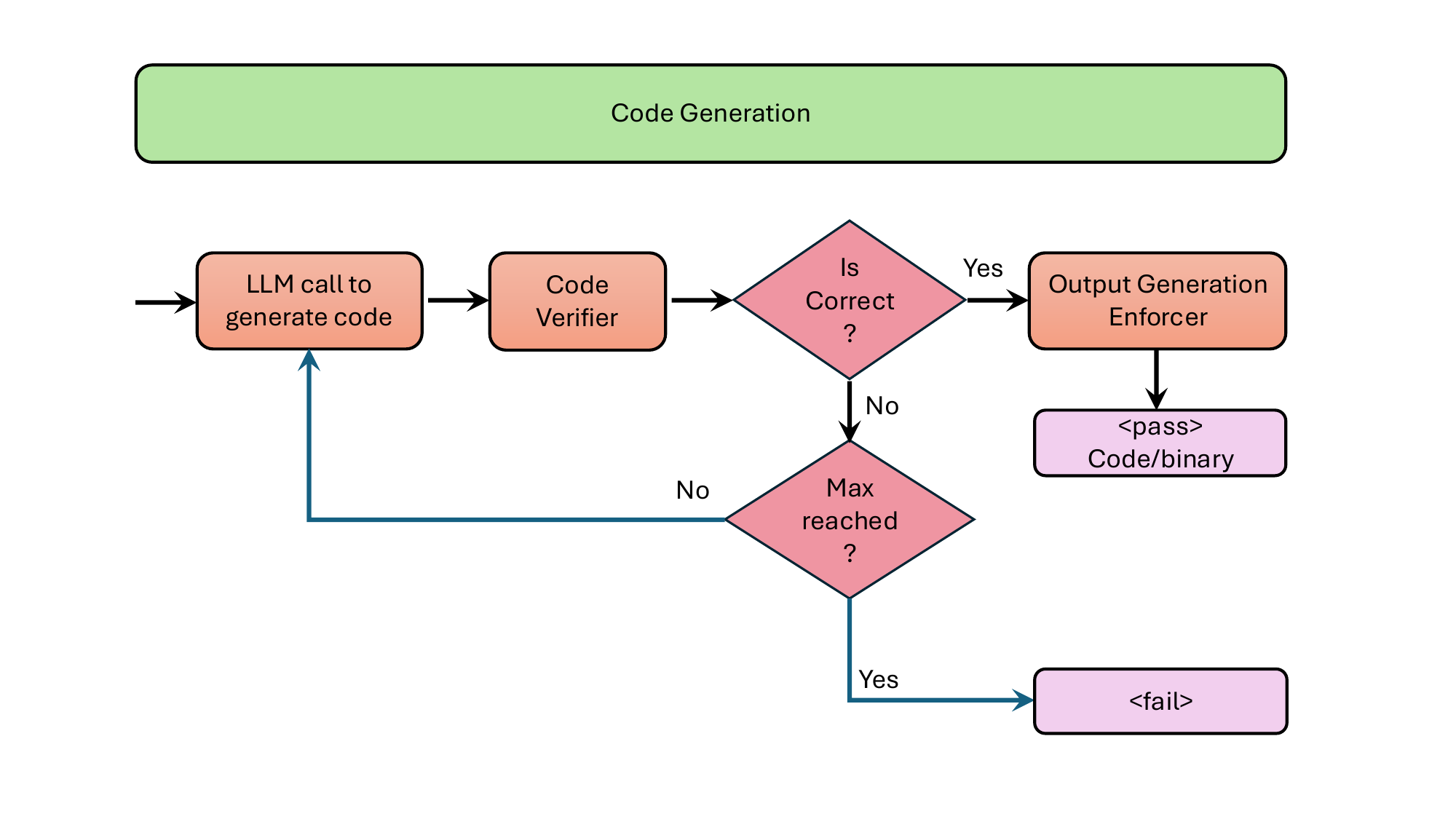}
    \caption{A description of the code generation algorithm. This is further discussed in Sec.~\ref{subsection: Step Interpretion Methods}.}
    \label{fig: code generation block diagram}
\end{figure*}

When either the step has the \texttt{(PYTHON)} prefix or the LLM Python-judge block decides that the step should be implemented using Python code, the Code Generator block (Fig.~\ref{fig: code generation block diagram}) and Code Executor block are called successively. The Code Generator block generates (Python is the preferred language in our implementation) and validates the code. After confirming that the code has a valid output generation instruction (i.e., it contains an instruction which provides the output required by the step), the block gives the $<$Success$>$ directive with the corresponding programming code in the call prompt. However, if this block fails to generate a valid output, it produces the $<$Fail$>$ directive as the call prompt. This output is then used by the Implement Step module to execute the step accordingly. Specifically, the code is generated using the GenerateCode function, which takes the task, instance and execution context as input and generates Python code to execute the step. However, we note that LLMs may fail to generate correct code. This includes syntax issues and runtime issues. To mitigate this issue, we use the CheckCode block, which utilizes an LLM to decide if the generated code is valid. This loop of code generation and verification continues until the code is deemed valid or the maximum number of iterations is exceeded. Even when the code is deemed correct, it still might not print the output of the code, which will later be captured by the subroutine. This is added using the CheckandAddPrint function. Finally, this code is used by Implement Step module and run by using the PythonCodeExecutor function. The step output obtained is finally added to the execution context. This is described in Algorithm~\ref{alg: Python Code Generator and Executor}.

\begin{algorithm}
\caption{Python Code Generator and Executor}
\label{alg: Python Code Generator and Executor}
\footnotesize
\begin{algorithmic}
    \STATE \textbf{Inputs:} Task $T$, Instance $I$, Execution Context $E$
    \STATE Initialize $\text{MAX\_ITERATIONS}$, $i = 0$
    \STATE Initialize $\text{is\_code\_valid}$ = "NO"
    
    \WHILE{\text{is\_code\_valid}=="NO" and $i<\text{MAX\_ITERATIONS}$}
        \STATE $\text{code} \gets$ GenerateCode$(T, I, E)$
        \STATE $\text{is\_code\_valid} \gets$ CheckCode(\text{code})
        \STATE $i \gets i+1$
    \ENDWHILE
    \STATE $\text{code} \gets$ CheckandAddPrint$(\text{code})$
    \STATE $\text{step\_output} \gets$ PythonCodeExecutor$(T,I,E,\text{code})$
    \STATE $E \gets E+\text{step\_output}$
\end{algorithmic}
\end{algorithm}

If a step is conditional, the condition is checked by the Interpret Step module and if the condition is met, the instruction (possibly having a prefix \texttt{(LLM)}, \texttt{(PYTHON)} or \texttt{(TOOL)}) is implemented accordingly. Otherwise, if the condition is not met the Interpret Step module gives the $<$``skip this step"$>$ directive to the Implement Step module.

Finally, if the instruction of the step has the prefix \texttt{(LLM)} or it is decided that a direct LLM call should be performed, the Interpret Step module produces the correct call prompt using the execution context, which is later used by the Implement Step Module.

\subsubsection{Implement Step Module}\label{subsection: Step Implementation Methods}
This module takes the call prompt of the Interpret Step module as input, and executes the step accordingly, i.e., by either (i) inputting the call prompt directly into an LLM, (ii) using the call prompt produced to call a tool to execute the code locally, or (iii) using the call prompt to call the generated code/binary locally and produce the output, or (iv) skipping the step. 

Next, the module checks the output in terms of its format, and also performs a semantic check with respect to the execution context. If this verification fails, the step is attempted again with the corresponding error rationale and the step implementation details from the execution context for a maximum number of attempts. If the step output passes verification, the step output is validated against constraints and rubrics (please see Section~\ref{subsection: Constraints}). If the constraints and rubrics checks fail, the step is executed again with the failed step's output and reason as feedback based on the context history until it passes, or a maximum number of trials is reached. If the maximum number of trials in the verification step or the constraints and rubrics checking steps is reached, we use a fall-back approach utilizing a Fall-Back LLM to perform the step by using the context history and move on without further verifications or checks. This fallback ensures that the step is implemented without syntax errors.

During execution, the executor maintains the execution context, implemented as an ordered message history, containing (i) the system specification and all previously accepted step instruction–output pairs which are also described in Section~\ref{subsection: Plan Implementation Representation}, and (ii) a result log which stores the details about each step in a readable format. In addition, the executor also maintains a running state summary which stores short summaries of the output for each step. This running summary is used for constraint validation at each step.

\begin{figure*}
    \centering
    \includegraphics[width=0.8\linewidth]{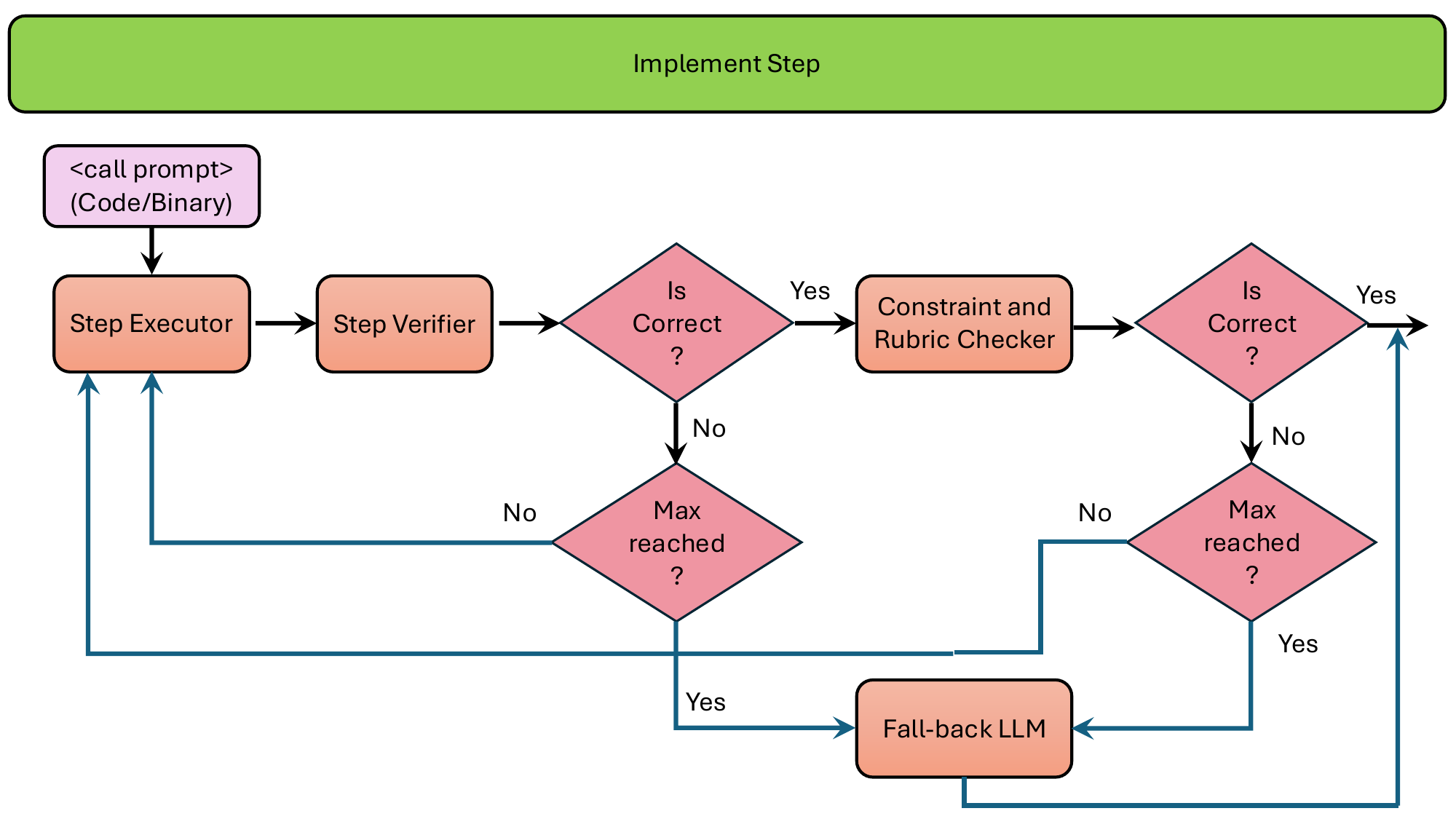}
    \caption{A description of the Implement Step module, which is a part of the Executor module. It is further described in Sec.~\ref{subsection: Step Implementation Methods}.}
    \label{fig: implement step block diagram}
\end{figure*}

The running state summary consists of short and factual descriptions of each accepted step's output. This summary is updated only when the new step's output is permanently appended to the execution context. This ensures that the running state summary contains only verified facts. The running state summary captures the current state of the plan while minimizing token usage and preventing unintended reasoning by downstream modules.

These features enable the executor to implement the plan as an interpretable program while supporting the constraint validation and error correction mechanisms described in Section~\ref{subsection: Constraints} and Section~\ref{subsection: Error Correction}.

\subsubsection{Constraints and Rubrics}\label{subsection: Constraints}
In Section~\ref{subsection: Tool Registry Setup and Constraint Generation}, we explained that in addition to the user-specified constraints and rubrics a set of constraints is also automatically generated based on the task and instance. All constraints and rubrics are aggregated and stored as strings containing a numbered list. We use these constraints to validate the output obtained from every step's implementation, if there is no implementation error. Before any validation is performed, the constraint list is parsed into a Python list for ease of use.

Once we have the step output of the current step, we first check whether there is a need to check constraints for the step using the ShouldCheckConstraintsForStep function. This is because many plan steps summarize information from the task, instance, and previous steps, and in these cases, the solution space is not modified by the step output. Therefore, there is no need to check if constraints are valid in this scenario. In such cases, we move directly to the next step's implementation.

If the function determines that the constraints should be checked for this step, then a subset of relevant constraints is first extracted by the FilterRelevantConstraints function. The purpose of this extraction is to filter out constraints which are not related to the current step, since checking them would inevitably lead to errors. The relevant constraints are then cached until the step output contains no errors and satisfies all relevant constraints, or the maximum number of retries for the step has been exhausted.

In RunAgent, each relevant constraint is checked against the step and the output of the step by a series of LLM calls. First, an LLM reasoner block is provided with the running state summary, the step output, and the relevant constraints to determine whether any constraint has been violated, along with a justification for its decision. Then, an LLM judge block reviews this reasoning to determine whether the constraint has been violated. This two-step process is used to avoid ambiguous responses from a single LLM call, which may sometimes indicate that a constraint is violated while providing reasoning that suggests otherwise. This two-step procedure is also employed because providing the task, instance, or execution context directly for validation can prompt the LLM to attempt solving the problem rather than strictly validating the current constraint. The neutral information-based nature of the running state summary forces the LLM to validate the current constraint. This is summarized in Algorithm~\ref{alg: Constraint Validation}.

\begin{algorithm}
\caption{Constraint Validation}
\label{alg: Constraint Validation}
\footnotesize
\begin{algorithmic}
    \STATE \textbf{Inputs:} Running State Summary $S$, Constraints $C$, Step instruction $s$, Step output $y$
    \IF{ShouldCheckConstraintsForStep(S,s,y)=="No"}
        \STATE \textbf{return}
    \ENDIF
    \STATE $C_{list}$ $\gets$ ParseConstraints($C$)
    \STATE $\text{Relevant\_}C_{List} = \text{FilterRelevantConstraints}(C_{List})$

    \STATE violated\_constraint\_list = []
    \FOR{constraint \textbf{in} Relevant\_$C_{List}$}
        \STATE reason = GiveReason(constraint, S, s, y)
        \STATE decision = ConstraintValid(reason)
        \IF{decision == "NO"}
            \STATE violated\_constraint\_list += [constraint, reason]
        \ENDIF
    \ENDFOR
    \STATE \textbf{return} violated\_constraint\_list
\end{algorithmic}
\end{algorithm}

\subsubsection{Error Correction}\label{subsection: Error Correction}
Whenever a step is implemented, there may be an error during implementation, which may arise from issues with LLM calls or during the implementation of Python code or tools. It is also possible that the step may be implemented without any such errors but the output does not satisfy the relevant constraints. In these cases, the step is implemented repeatedly until the output is valid and satisfies all the relevant constraints. 

To prevent RunAgent from repeating previously encountered errors, the cause of an error is temporarily appended to the execution context using the XML tags \texttt{<error>} and \texttt{</error>} when the error occurs during Python code, tool, or LLM execution. If a subset of constraints is violated, the violated constraints, the rationale for the violations, and the corresponding output are appended as a structured error message. Moreover, the LLM prompts in the execution context are modified to explicitly inform the model of previously encountered errors, thereby conditioning subsequent attempts on a clear description of earlier failures.

Once all errors and constraint violations are resolved, the temporarily appended error and violation information is removed from the execution context, and only the step output is appended in the manner specified in Section~\ref{subsection: Plan Implementation Representation}.

If errors or constraint violations persist after the maximum number of retries, RunAgent falls back to direct Fall-Back LLM-based step execution using the execution context to ensure continued execution. This is described in Algorithm~\ref{alg:runagent_error_correction_simple}.

\begin{algorithm}[t]
\caption{Error Correction Implementation}
\label{alg:runagent_error_correction_simple}
\footnotesize
\begin{algorithmic}[1]
\STATE \textbf{Inputs:} Plan step $s$, execution context $E$, $\text{MAX\_RETRIES}$
\STATE $i \gets 0$

\WHILE{$i \leq \text{MAX\_RETRIES}$}
    \STATE $y \gets$ ImplementStep$(s, E)$
    \IF{Error$(y)$}
        \STATE $err \gets$ MakeImplementationErrorMessage$(s,y,E)$
        \STATE $E \gets E \,+\, \langle error \rangle err \langle /error \rangle$
        \STATE $i \gets i + 1$
        \STATE \textbf{continue}
    \ENDIF
    \IF{ViolatesConstraints$(s, y)$}
      \STATE $err \gets$ MakeConstraintErrorMessage$(s, y,E)$ 
      \STATE $E \gets E \,+\, \langle error \rangle err \langle /error \rangle$
      \STATE $i \gets i + 1$
      \STATE \textbf{continue}
    \ENDIF
    \STATE $E \gets$ RemoveTemporaryErrors$(E)$
    \STATE $E \gets$ AppendStepOutput$(E, s, y)$
\ENDWHILE

\IF{$i>\text{MAX\_RETRIES}$}
    \STATE $y \gets$ FallbackLLM$(s, E)$
    \STATE $E \gets$ AppendStepOutput$(E, s, y)$
\ENDIF
\STATE \textbf{return} $E$
\end{algorithmic}
\end{algorithm}

\subsubsection{Facts Utilization}
Facts represent a mechanism for users to interact with RunAgent by providing immutable facts or suggestions as auxiliary information which might be implicit or not mentioned in the task-instance pair or the plan. A user may provide such facts either prior to execution or as feedback for subsequent attempts when the plan implementation does not meet expectations. By reviewing the plan execution or logs, the user can specify corrective information, enabling RunAgent to adjust its behavior and implement the plan correctly in the next iteration.

In some cases, the facts may provide a particular interpretation of the meaning or clarification of a concept. For example, a ``day" may sometimes refer to a full day including the daytime and nighttime, while in some other context, it may only refer to daytime. In some cases, a day could mean a full 24-hour day, and in some other contexts, a partial day may constitute a day. RunAgent mainly utilizes the facts in the Constraints and Rubrics Checking mechanism to ensure that the facts in the view of the context are followed. In particular, (i) the facts are used to determine whether constraints should be evaluated in the first place, and (ii) the facts are used during constraint checking to determine whether the current step’s output satisfies the relevant constraints. It is important to note that these facts do not directly influence RunAgent’s execution, but instead serve solely to assist in constraint validation. Some examples of Facts are given in the Appendix.

\section{Evaluation}\label{sec:evaluation}

\subsection{Experimental Setup}
To evaluate the quality of plan implementation with RunAgent, we utilize two datasets from Natural-plan, i.e., Calendar Scheduling and Trip Planning datasets \cite{zheng2024natural}, and three datasets from SciBench, i.e, Stat, Calc, and Diff datasets \cite{wang2024SciBench}. All LLM calls and LLM blocks in RunAgent in this paper utilize GPT-4o \cite{hurst2024gpt}. We compare our results with the vanilla GPT-4o as a baseline to either directly solve the problem or implement a plan. We do not evaluate the Meeting Scheduling dataset from Natural-plan due to the existence of many correct answers for each question in the dataset. We use the SciBench Math subset in our evaluation since the subset does not contain images with the exception of one question from the Calc dataset that is not included in the evaluation. We also compare our results for the Natural-plan Calendar Scheduling and Natural-plan Trip Planning datasets with the state-of-the-art PlanGEN algorithms \cite{parmar2025plangen}. All results for Gemini-1.5-Pro and Gemini-2.0-Flash are taken from \cite{parmar2025plangen}.

We use accuracy as a measure of performance for the SciBench Math dataset and the Exact Match (EM) accuracy as a measure of performance for the Natural-plan Calendar Scheduling and Natural-plan Trip Planning datasets where multiple correct answers may be viable in some cases. We use an LLM-as-a-judge to check if the produced answer is equivalent to the Gold Answer from the dataset. Finally, because RunAgent implements plans, we use a custom algorithm (which uses GPT-4o internally to generate plans) to generate feasible plans which are then implemented by RunAgent. In order to have a fair comparison, we also implement the custom plans we generate using GPT-4o, and report the accuracies under `GPT-4o Plan Implementation'. Since plan generation is not the focus of this paper, the details are omitted.

\begin{table}[t]
    \centering
    \small
    \begin{tabular}{|c|c|}
    \hline
    \textbf{Method}  &  \textbf{EM Acc.} ($\%$)\\
    \hline
    GPT-4o Plan Implementation & 46.9\\
    Gemini-1.5-Pro & 48.9\\
    GPT-4o & 58.3\\
    PlanGEN, Best of N, Gemini-1.5-Pro & 60.7\\
    Gemini-2.0-Flash & 61.1\\
    PlanGEN, Best of N, Gemini-2.0-Flash & 68.9\\
    RunAgent without Constraint Checking & 75.4\\
    RunAgent & 81.1\\
    \hline
    \end{tabular}
    \caption{\textbf{Results for Natural-plan Calendar Scheduling Dataset}}
    \label{tab:calendar results}
\end{table}

\subsection{Main Results}

\noindent\textbf{Performance on Natural-plan Datasets}: The evaluation results for the Calendar Scheduling dataset are summarized in Table~\ref{tab:calendar results}. PlanGEN BoN (best-of-N) \cite{parmar2025plangen} as a State-of-the-art planner and executor achieves significant improvement over the respective baselines of Gemini-Flash. RunAgent considerably improves the accuracy beyond PlanGEN BoN despite internally using GPT-4o whose baseline performance is situated between Gemini-1.5-flash and Gemini-2.0-flash. We also perform an ablation study by deactivating run-time constraint checking by RunAgent and we observe that the performance drops from $81.1\%$ to $75.4\%$ and yet achieving superior performance over PlanGEN BoN.

\begin{figure*}[t]
    \centering
    \includegraphics[width=\linewidth]{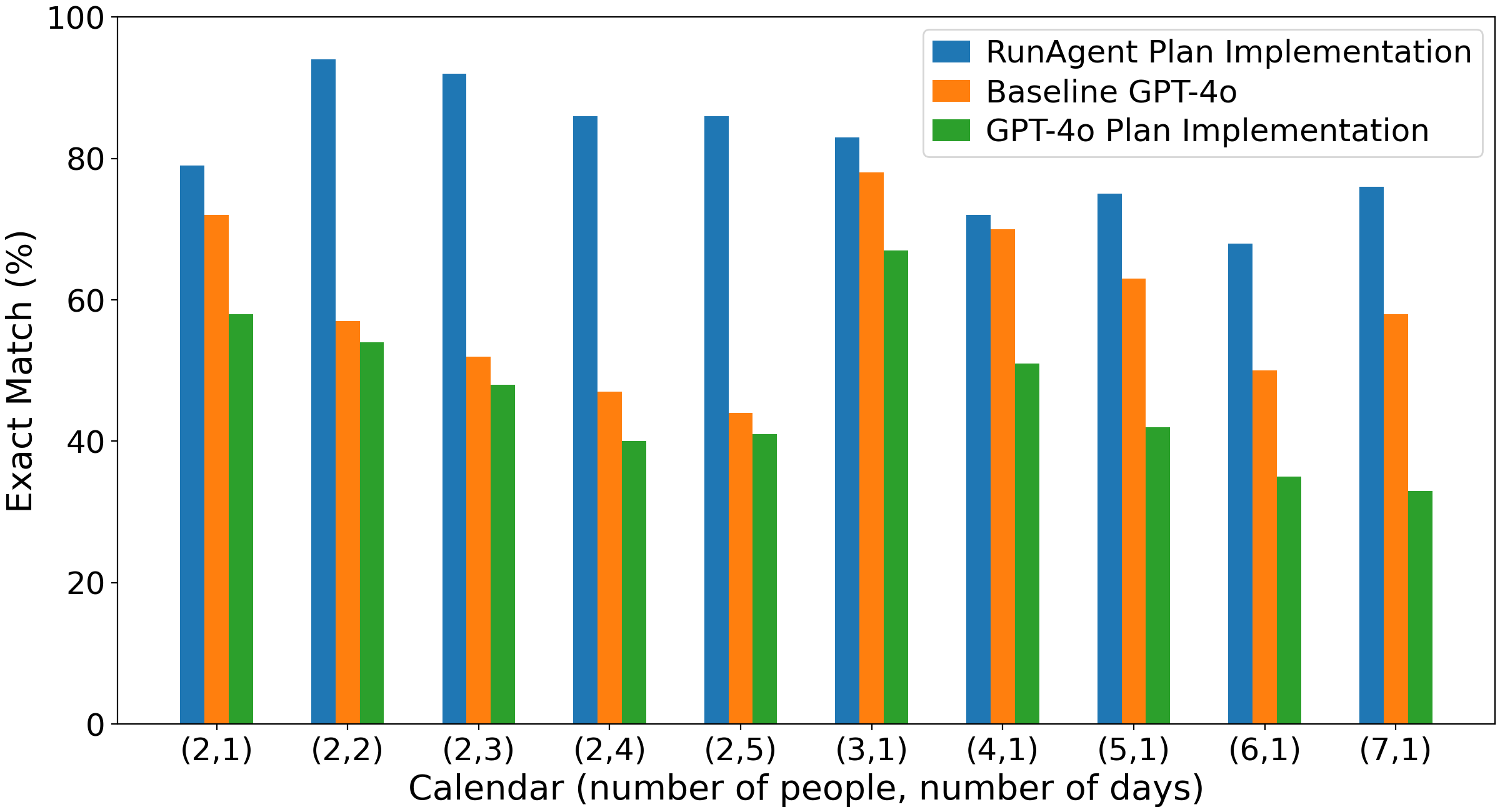}
    \caption{Accuracy for different problem complexities (Calendar Scheduling)}
    \label{fig: calendar results}
\end{figure*}

Finally, we analyze RunAgent performance for different levels of problem difficulty, which is described in Fig.~\ref{fig: calendar results}. The Calendar Scheduling dataset is partitioned into 10 bins of increasing difficulty, each containing 100 problems with fixed numbers of people and days. We observe that RunAgent consistently performs well in all bins with overall slight degradation in performance as the difficulty of the problems increases. Aside from minor performance fluctuations, we observe a notable drop in the first bin (two people, one day), which we attribute to the presence of multiple valid solutions in this setting. As a result, although RunAgent often produces correct solutions, they may not exactly match the Gold Answers. Upon manual inspection of the 188 mismatched cases in all bins, we find that 51 are verified as correct, yielding an adjusted accuracy of $86.2\%$ instead of $81.1\%$. The evaluation results for the Trip Planning dataset are presented in Table~\ref{tab:trip results} which show a similar trend as Calendar Scheduling dataset.

\noindent\textbf{Performance on SciBench Math Dataset}: The evaluation results for Calc, Stat, and Diff datasets can be seen in Table~\ref{tab:SciBench results}. It is observed that RunAgent considerably improves the accuracy over its baseline LLM, i.e., GPT-4o. However, planning is not enough without an effective execution platform. Indeed, the performance slightly drops from that of the baseline GPT-4o, when GPT-4o itself is instructed to solve a task and instance by implementing the same plan used by RunAgent. A similar trend is observed for the Calendar Scheduling dataset.

\noindent\textbf{Discussion}: We observe that GPT-4o performs better when solving the problem directly than when guided by a feasible plan. This occurs because plan-based prompting shifts the model’s focus from internal problem-solving to plan execution, which can hinder its reasoning process, since it does not guarantee step-by-step execution, as GPT-4o may interpret plans holistically. In contrast, RunAgent enforces stepwise execution with integrated verification, constraint checking, and reasoning, enabling superior performance.

\begin{table}[t]
    \centering
    \small
    \begin{tabular}{|c|c|}
    \hline
    \textbf{Method}  &  \textbf{EM Acc.} ($\%$)\\
    \hline
    GPT-4o & 3.07\\
    GPT-4o Plan Implementation & 6.69\\
    RunAgent & 14.73\\
    Gemini-1.5-Pro & 34.75\\
    PlanGEN, Best of N, Gemini-1.5-Pro & 41.63\\
    \hline
    \end{tabular}
    \caption{\textbf{Results for Natural-plan Trip Planning Dataset}}
    \label{tab:trip results}
\end{table}

\begin{table}[t]
    \centering
    \small
    \begin{tabular}{|c|c|c|c|}
    \hline
    \textbf{Method}  &  Stat & Calc & Diff\\
    \hline
    GPT-4o & 72.22 & 70.73 & 50\\
    GPT-4o Plan Implementation & 70.83 & 65.85 & 42\\
    RunAgent & 80.56 & 78.05 & 62\\
    \hline
    \end{tabular}
    \caption{\textbf{Accuracy Percentage for the SciBench Dataset}}
    \label{tab:SciBench results}
\end{table}

\section{Conclusion}\label{sec:conc}
While strategic planning is central to problem-solving and AI agent development, its effectiveness critically depends on both an appropriate language for plan representation, and robust plan execution. RunAgent provides an initial solution that highlights the importance of these two factors.

\bibliographystyle{unsrt}
\bibliography{refs}

\appendix

\subsection{Prompts}

\subsubsection{Initialization Prompts}

\begin{tcolorbox}[colback=gray!10, colframe=black, title=Constraint Generation Prompt]
\footnotesize messages=[
            \{"role": "system", "content": """You are an expert at extracting constraints that MUST NOT be violated when solving a task instance.

You will be given:\newline
- A task description\newline
- A specific instance of the task\newline

Your job:\newline
- Output a comprehensive list of constraints that should not be violated while solving the instance.\newline
- Include all explicit hard requirements from the task/instance.\newline
- Include any implicit but necessary constraints/assumptions needed to correctly interpret the instance.\newline
- Do not include any solution, schedule, or plan, ONLY constraints.\newline

Critical constraint quality requirements:\newline
- Make constraints ATOMIC: each numbered item must contain exactly ONE requirement.\newline
- If a requirement contains "and", "or", multiple clauses, or multiple checks, split it into multiple numbered constraints.\newline
- Avoid compound / multi-part constraints, even if the original text is compound, decompose it.\newline
- Avoid solver goals and optimization goals (e.g., "find the best", "maximize", "minimize"). Only hard validity/legality constraints.\newline

Output format requirements (critical):
- Output ONLY a numbered list of constraints as plain text (e.g., '1....\\n2. ...').\newline
- No headers, no explanations, no JSON, no code fences, no extra text.
"""\},\newline
            \{"role": "user", "content": """Task:\{task\}\newline

Instance:
\{instance\}
"""\},
        ]
\end{tcolorbox}

\subsubsection{Python Code Execution Prompts}

\begin{tcolorbox}[colback=gray!10, colframe=black, title=Python Code Add Print System Prompt]
    \footnotesize You are an expert at recognizing and adding print statements to Python code. You are given a Python code \{code generated\}. If the code prints an output, output "No" without any other text. If the code does not print an output, add an appropriate print statement to the code to print an output and return the modified code. Do not add any other text to the output.
\end{tcolorbox}

\begin{tcolorbox}[colback=gray!10, colframe=black, title=Python Code Generation System Prompt]
    \footnotesize Let's think step by step. You are given a prompt, which contains context and a step that needs to be carried out at the end. \newline Write a detailed code in Python that will carry out the step. The code should use the standard libraries of Python.\newline Use of sympy, scipy or numpy is also allowed.\newline The result of the code implementation should be printed using the Python print function.\newline Do NOT catch exceptions in your code; let any errors be unhandled.\newline The output should contain only a string of Python code, and the arguments required for the code. The input text is: \{instance\}
\end{tcolorbox}

\begin{tcolorbox}[colback=gray!10, colframe=black, title=Python Code Checker System Prompt]
    \footnotesize You are an expert at recognizing Python code. You are given an input \{code generated\}. If the input is valid Python code, output "Yes". Else output "No". Do not add any other text to the output.
\end{tcolorbox}

\subsubsection{Constraint Validation Prompts}

\begin{tcolorbox}[colback=gray!10, colframe=black, title=Python Code Execution System Prompt]
\footnotesize You are given a conversation, where at the end of the conversation a step is supposed to be carried out. \newline The Python code to implement the last step is also given. Execute the Python code.\newline If you reply with a Function Call, you must carefully provide the arguments required to run the Python code by considering the input types for the input in the function description.\newline The arguments you provide must not contain any forward slashes.
\end{tcolorbox}

\begin{tcolorbox}[colback=gray!10, colframe=black, title=Restriction Decision Prompt]
\footnotesize messages=[
            \{"role": "system", "content": """You are an expert at classifying whether a single step in plan execution is restricting the solution space.\newline

You will be given:\newline
- Task description\newline
- Instance of the task\newline
- The CURRENT step (instruction and output)\newline

Definitions:\newline
- "Restricting" means the step output makes a commitment/decision that narrows possibilities (e.g., assigning values, choosing an option, fixing an allocation, ruling out options, etc.).\newline
- "Not restricting" means the step output only records facts, extracts information from the input or lists candidates/options.\newline

Output format (CRITICAL):\newline
Output ONLY 'Yes' or 'No'. Do not output anything else.
Facts: \{facts\}
"""
\},\newline
\{"role": "user", "content": """Task: \{task\}\newline
Instance: \{instance\}\newline
CURRENT STEP (instruction + output):\{step\_text\}
"""\},
        ]
\end{tcolorbox}

\begin{tcolorbox}[colback=gray!10, colframe=black, title=Constraint Relevance Prompt]
\footnotesize messages = [\{
            "role": "system",
            "content": f"""You are an expert in checking if a given constraint is necessary and relevant for a specific step in plan execution. You are given a step in an ongoing plan execution, along with a constraint that needs to be checked.
Your job is to determine if this constraint is necessary and relevant and should be verified for this specific step. This means that the constraint should be included if the step output can violate it.\newline

Task description: \{task\}\newline
Instance of the task: \{instance\}\newline
Current step number: \{step number\}\newline
Current step instruction: \{step instruction\}\newline
Constraint to evaluate: \{constraint\}\newline

Consider:\newline
- Does this constraint apply to the output or behavior expected from this specific step?\newline
- Is it meaningful to check this constraint after this step completes?\newline
- Is it necessary to check this constraint for this step?\newline
- Would checking this constraint help ensure the step was executed correctly?\newline

Only output 'Yes' if the constraint should be necessarily checked for this step, or 'No' if it is not relevant or necessary to check for this step. Do not output anything else."""
        \},\newline
        
        \{"role": "user", "content": f"Should the constraint '\{constraint\}' be necessarily checked for this step? Answer only 'Yes' or 'No'. If the constraint is not relevant or necessary to check for this step, answer 'No'."\}]
        
\end{tcolorbox}

\begin{tcolorbox}[colback=gray!10, colframe=black, title=Constraint Checker Reason Generation Prompt]
\footnotesize messages=[
            \{"role": "system", "content": """You are an expert in checking if a given constraint is violated by the given step's output during plan execution. You are not a solver, and you do not care if the plan succeeds or fails. You only care if it violates the constraints.\newline
            Given the conversation so far and a constraint, your job is to check if the given last executed step's output violates the constraint from a human's perspective.\newline
            It is critical that you apply human intuition and common sense to check if the constraint is violated. Think of the constraint being set in the real world and the step's output being the action taken by a human.\newline
            In doing so, you will look for the following:\newline
            - Does the step involve inspection of the problem or does it perform an action which makes a decision that needs to be verified?\newline
            - If the step inspects the problem, such as gathering information, listing options, checking status etc., then the constraint is not violated and you should output "No", followed by this reason.\newline
            - If the step performs an action which restricts the solution space, then compare the output of the step with the constraint and output "Yes" if the constraint is violated and "No" if the constraint is not violated followed by the reason.
            DO NOT TRY TO SOLVE THE PROBLEM, ONLY CHECK IF THE STEP OUTPUT IS LEGAL UNDER THE GIVEN CONSTRAINT.
            Ignore any special characters in the last executed step's output.\newline
            \{facts\}\{current state\}\newline
            Output ONLY a detailed sentence/paragraph explaining whether the constraint is violated or not, and why.\newline
            Do NOT output JSON. Do NOT output 'Yes' or 'No' labels. Print ONLY the explanation paragraph."""\},\newline
            \{"role": "user", "content": """Check if the following constraint is violated by the last executed step's output: \{constraint\}
"""\},
        ]
\end{tcolorbox}

\begin{tcolorbox}[colback=gray!10, colframe=black, title=Constraint Checker Decision Prompt]
\footnotesize messages=[
            \{"role": "system", "content": """You are a strict judge. You will be given:\newline
- A constraint\newline
- A detailed reason describing whether the constraint is violated or not\newline

Your job is to output ONLY:\newline
Yes  (if the constraint is violated)\newline
No   (if the constraint is not violated)\newline

Do not output anything else."""\},\newline
            \{"role": "user", "content": """Constraint:\{constraint\}\newline
Reason:\{reason\}
"""\},
        ]
\end{tcolorbox}

\begin{tcolorbox}[colback=gray!10, colframe=black, title=State Summarizer Prompt]
\footnotesize messages=[
            \{"role": "system", "content": """You are a state summarizer used ONLY for constraint verification.\newline

You will be given:\newline
- The instruction for the most recent step\newline
- The output produced by that step\newline

Your job:\newline
- Summarize ONLY the key factual results from this (instruction, output) pair.\newline
- Do NOT propose next actions, alternatives, improvements, or any solution.\newline
- Do NOT infer facts that are not explicitly present in the output.\newline
- If the output is empty/unclear, say so.\newline

Output requirements:\newline
- Output plain text only.\newline
- No JSON, no code fences, no headers.\newline
"""\},\newline
            \{"role": "user", "content": """STEP INSTRUCTION:\{step instruction\}\newline
STEP OUTPUT:\{step output\}
"""\},
        ]
\end{tcolorbox}

\subsubsection{Executor Prompts}

\begin{tcolorbox}[colback=gray!10, colframe=black, title=Executor System Prompt]
\footnotesize You will execute a series of steps on a given input.\newline Task: \{task\}\newline Instance: \{instance\}.\newline
    You may be provided a preamble before the steps. Each step may have constraints. \newline

    Every time I ask you to execute a step, you will generate just the exact output for the step without any further explanation. 
    The message with the step will be in the following format:
    <step><number>number of the step$<$/number$><$instruction$>$instruction for the step$<$/instruction$><$/step$>$\newline

    The output of the previous steps will be in the following format:
    $<$step$><$number$>$number of the step$<$/number$><$output$>$output for the step$<$/output$><$/step$>$\newline
    Every time you output a step, you will output the entire step in the above format.\newline
    You will never output the next step instructions as an output of a step.\newline
    If a step requires you to take an input from a previous step, you will locate that previous step's output using the number of the step and output identifier and use it as the input for the current step.\newline

    \{function descriptions\}
\end{tcolorbox}

\subsubsection{Compiler Prompts}
\begin{tcolorbox}[colback=gray!10, colframe=black, title=Python Judge System Prompt]
\footnotesize You are asked to judge whether a step at the end of a conversation can be implemented better as a Python code or by LLM API call. \newline
            You know that LLMs are good at problems where natural language processing and creativity are required.
            You know that LLMs are not good at problems where algorithmic thinking and mathematical reasoning are required.
            You know that Python codes are good at problems where algorithmic thinking and mathematical reasoning are required.
            You know that Python codes are not good at problems where natural language processing and creativity are required.
            You are given a conversation in which a plan is being implemented. The final step of the input from the user is to be implemented.\newline
            You will reply with strictly one word 'Yes' or 'No' answer to the question: Should the final step be implemented in Python?\newline
            If the answer is 'Yes', output 'Yes'.\newline
            If the answer is 'No', output 'No'.\newline
            Do not output anything else.\newline
\end{tcolorbox}

\subsubsection{Keyword Prompts}
\begin{tcolorbox}[colback=gray!10, colframe=black, title=FORALL Prompt]
\footnotesize messages = [\{"role": "system", "content": """You are given a FORALL statement from the user. Below you are given the associated task, an instance and a chat history. \newline
The task is: \{task\}.\newline
The instance is: \{instance\}.\newline
The chat history is: \{chat history\}.\newline
Your job is to use the task, instance and chat history to find all items that the FORALL statement iterates over and the task that needs to be executed for each item.\newline
The output should be as a dictionary in Python: \{\{"items": [item1, item2, item3, ...], "task": "task that needs to be executed for each item"\}\}"""\},

\{"role": "user", "content": \{step instruction\}\}]
\end{tcolorbox}

\begin{tcolorbox}[colback=gray!10, colframe=black, title=IF Prompt]
\footnotesize messages = [\{"role": "system", "content": """You are given an if statement. Your task is to separate the if statement into the condition and the then statement.\newline
If the statement tells you to go to a step, the statement should be in the following format: goto step $<$step number$>$.\newline
The output should be in the following dictionary format:\newline
\{"if condition": "condition","then statement": "statement"\}"""\},

\{"role": "user", "content": \{step instruction\}\}]
\end{tcolorbox}

\subsection{Examples of Keyword Outputs}

\subsubsection{Example for FORALL}

\paragraph{Task}
Write short creative stories about New York monuments. Execute the numbered plan exactly. Every story should be about 100 words and focused on the monument named in the iteration.
\paragraph{Input}
\begin{quote}
New York monuments and landmarks to feature: Statue of Liberty, Brooklyn Bridge, Empire State Building.
\end{quote}
\paragraph{Plan and Observed Output}
\begin{description}
\item
\textbf{Plan:} 1. FORALL New York monuments or landmarks listed in the input text: LLM: Write a vivid story of about 100 words centered on that monument. \\
\textbf{Observed output for Statue of Liberty:} ``At dawn, the Statue of Liberty stands silent, her torch cutting through the gentle mist like a lighthouse guiding weary souls. Once a welcoming beacon for immigrants arriving by sea, her green copper face reflects the dreams of those seeking freedom. Emma, a young artist, sketches the statue, feeling the weight of history in every stroke. The air hums with stories of hope and resilience, whispered by the harbor breeze. As the sun rises, Lady Liberty's torch gleams brightly, a timeless emblem of courage and liberty shining over New York, inspiring all who behold her to embrace new beginnings.'' \\
\textbf{Observed output for Brooklyn Bridge:} ``As dawn breaks over Manhattan, the Brooklyn Bridge stands majestic, its cables woven like a giant harp strummed by the city’s heartbeat. Emma, clutching her grandfather’s old camera, steps onto the wooden walkway, the mist curling around steel towers like whispered legends. Each footstep echoes stories of dreamers, engineers, and workers who defied odds to build this marvel. Below, the East River glistens, ferrying hopes and histories. Emma raises her camera, capturing the sun’s golden kiss on the Gothic arches. In this moment, the bridge connects past and present, her memories intertwining with the city’s timeless song.'' \\
\textbf{Observed output for Empire State Building:} ``In the heart of New York City, the Empire State Building stands as a sentinel of dreams and determination. As twilight paints the sky in hues of orange and pink, the skyscraper’s lights flicker to life, a beacon for dreamers worldwide. Long ago, ambitious architects and workers toiled through freezing winters and scorching summers, crafting this marvel from steel and stone. Beneath its towering spire, lovers meet, tourists gaze in awe, and New Yorkers find a moment of solace amid the city's chaos. The Empire State Building is more than concrete; it’s the spirit of ambition etched into the skyline, forever inspiring.''
\end{description}

\subsubsection{Example for IF (True)}
\paragraph{Task}
Demonstrate conditional execution with simple mathematical statements. Execute the numbered plan exactly, and when a step asks for exact text, reproduce it verbatim with no extra words.
\paragraph{Input}
\begin{quote}
Math note: 12 is greater than 7, 3 is less than 10, and no arithmetic errors are present.
\end{quote}
\paragraph{Plan and Observed Output}
\begin{description}
\item[Step 1]
IF 12 is greater than 7 according to the input text, then LLM: Output exactly ``Math branch activated because 12 $>$ 7.'' \\
\textbf{Observed output:} ``Math branch activated because 12 $>$ 7.''
\item[Step 2]
LLM: Output exactly ``Continue with the verified arithmetic storyline.'' \\
\textbf{Observed output:} ``Continue with the verified arithmetic storyline.''
\item[Step 3]
LLM: Output exactly ``Mathematical IF demonstration complete.'' \\
\textbf{Observed output:} ``Mathematical IF demonstration complete.''
\end{description}

\subsubsection{Example for IF (False)}
\paragraph{Task}
Demonstrate conditional execution with simple mathematical statements. Execute the numbered plan exactly, and when a step asks for exact text, reproduce it verbatim with no extra words.
\paragraph{Input}
\begin{quote}
Math note: 5 is less than 9, 2 plus 2 equals 4, and 10 is not less than 3.
\end{quote}
\paragraph{Plan and Observed Output}
\begin{description}
\item[Step 1]
IF 10 is less than 3 according to the input text, then LLM: Output exactly ``False math branch activated.'' \\
\textbf{Observed output:} no step output was produced; the branch was skipped.
\item[Step 2]
LLM: Output exactly ``Continue after the false mathematical IF.'' \\
\textbf{Observed output:} ``Continue after the false mathematical IF.''
\item[Step 3]
LLM: Output exactly ``False mathematical IF demonstration complete.'' \\
\textbf{Observed output:} ``False mathematical IF demonstration complete.''
\end{description}

\subsubsection{Example for GOTO}
\paragraph{Task}
Demonstrate direct control-flow jumps. Execute the numbered plan exactly, and when a step asks for exact text, reproduce it verbatim with no extra words.
\paragraph{Input}
\begin{quote}
Routing note: jump directly to the emergency branch and skip the descriptive intermediate steps.
\end{quote}
\paragraph{Plan and Observed Output}
\begin{description}
\item[Step 1]
goto step 4 \\
\textbf{Observed output:} no step output was recorded.
\item[Step 2]
LLM: Output exactly ``This step should be skipped.'' \\
\textbf{Observed output:} no step output was recorded.
\item[Step 3]
LLM: Output exactly ``This step should also be skipped.'' \\
\textbf{Observed output:} no step output was recorded.
\item[Step 4]
LLM: Output exactly ``Reached the later step through direct GOTO.'' \\
\textbf{Observed output:} ``Reached the later step through direct GOTO.''.
\item[Step 5]
LLM: Output exactly ``Direct GOTO demonstration complete.'' \\
\textbf{Observed output:} ``Direct GOTO demonstration complete.''.
\end{description}

\subsection{Example Outputs of Executed Plans}

\subsubsection{Example for Natural Plan Calendar Scheduling}
\begin{tcolorbox}[colback=gray!10, colframe=black, title=Executor System Prompt]

\footnotesize Task: Your task is to find a feasible schedule for calendar events, taking into account time slots, durations, constraints, and conflicts. You will analyze the calendar requirements and available time slots to determine the optimal event schedule. Output the final calendar schedule outlining the day and time of the meeting, without any other text.

Instance:
 You need to schedule a meeting for Michelle, Steven and Jerry for one hour between the work hours of 9:00 to 17:00 on Monday. 

Here are the existing schedules for everyone during the day: 
Michelle has meetings on Monday during 11:00 to 12:00; 
Steven has blocked their calendar on Monday during 9:00 to 9:30, 11:30 to 12:00, 13:30 to 14:00, 15:30 to 16:00; 
Jerry has blocked their calendar on Monday during 9:00 to 9:30, 10:00 to 11:00, 11:30 to 12:30, 13:00 to 14:30, 15:30 to 16:00, 16:30 to 17:00; 

Find a time that works for everyone's schedule and constraints. 
SOLUTION: 

Plan Implementation: [
  \{\newline
    "step\_number": "1",\newline
    "step": "Determine the available times for each participant by examining the end times of busy slots, as participants are considered free at those times.",\newline
    "reasoner": "Tools: Python Executor",\newline
    "output": "Michelle is available at the following times: [9, 12, 17]\newline
    Steven is available at the following times: [9, 9.5, 12, 14, 16, 17]\newline 
    Jerry is available at the following times: [9, 9.5, 11, 12.5, 14.5, 16, 17, 17]",\newline
    "error": null\newline
  \},\newline
  \{\newline
    "step\_number": "2",\newline
    "step": "Cross-check all calendar events that need to be scheduled, taking note of their durations and any constraints or conflicts, to identify overlapping free time slots for all participants.",\newline
    "reasoner": "LLM",\newline
    "output": "$<$step$>$$<$number$>$2$<$/number$><$output$>$Available common time slots for all participants: [12:30 - 13:00], [14:30 - 15:30]$<$/output$><$/step$>$",\newline
    "error": null\newline
  \},\newline
  \{\newline
    "step\_number": "3",\newline
    "step": "Evaluate each event to determine its scheduling requirements, such as specific time constraints, dependencies on other events, and match them to the common free time slot that satisfies the constraints.",\newline
    "reasoner": "LLM",\newline
    "output": "$<$step$><$number$>$3$<$/number$><$output$>$The meeting can be scheduled from 14:30 to 15:30, as it fits the one-hour duration requirement and maintains alignment with all participants' availability.$<$/output$><$/step$>$",\newline
    "error": null\newline
  \},\newline
  \{\newline
    "step\_number": "4",\newline
    "step": "Prioritize the common time slots based on urgency or strategic importance, ensuring that finding a shared time slot for key events is prioritized.",\newline
    "reasoner": "LLM",\newline
    "output": "$<$step$><$number$>$4$<$/number$><$output$>$The meeting is prioritized to be scheduled from 14:30 to 15:30 due to the one-hour requirement and aligned schedules with all participants, with no further urgency or strategic importance evaluations needed.$<$/output$><$/step$>$",\newline
    "error": null\newline
  \},\newline
\end{tcolorbox}
\begin{tcolorbox}[colback=gray!10, colframe=black, title=]

\footnotesize
\{\newline
    "step\_number": "5",\newline
    "step": "Select a suitable one-hour time slot shared by all participants, matching the duration and constraints of the highest priority event first.",\newline
    "reasoner": "Tools: Python Executor",\newline
    "output": "14:30 to 15:30",\newline
    "error": null\newline
  \},\newline
  \{\newline
    "step\_number": "6",\newline
    "step": "Adjust the selected time slot, utilizing free slots identified, to resolve any conflicts or unmet constraints among participants.",\newline
    "reasoner": "LLM",\newline
    "output": "$<$step$><$number$>$6$<$/number$><$output$>$No further adjustments are needed for the selected time slot of 14:30 to 15:30, as it meets all constraints and does not conflict with any participants' schedules.$<$/output$><$/step$>$",\newline
    "error": null\newline
  \},\newline
\{
    "step\_number": "7",\newline
    "step": "Review the proposed time slot to ensure that it satisfies all participant availability and event constraints.",\newline
    "reasoner": "LLM",\newline
    "output": "$<$step$><$number$>$7$<$/number$><$output$>$The proposed time slot of 14:30 to 15:30 satisfies all participant availability and event constraints, ensuring a conflict-free schedule for the meeting.$<$/output$><$/step$>$",\newline
    "error": null\newline
  \},\newline
  \{
    "step\_number": "8",\newline
    "step": "Output the finalized schedule, detailing the day and time of the event according to the requirements, ensuring all participants share the chosen slot.",\newline
    "reasoner": "LLM",\newline
    "output": "$<$step$><$number$>$8$<$/number$><$output$>$Monday, 14:30 to 15:30$<$/output$><$/step$>$",\newline
    "error": null\newline
  \},\newline
  \{\newline
    "step\_number": "9",\newline
    "step": "FIN",\newline
    "output": "FINISHED",\newline
    "error": null\newline
  \}\newline
]

\end{tcolorbox}

\subsubsection{Example for Scibench Stat}
\begin{tcolorbox}[colback=gray!10, colframe=black, title=]
\footnotesize Task: Your task is to solve statistics and probability problems by analyzing the problem statement, applying appropriate statistical methods, and calculating the required values. Output the final answer in the required format, without any other text.\newline

Instance:
If the distribution of $Y$ is $b(n, 0.25)$, give a lower bound for $P(|Y / n-0.25|<0.05)$ when $n=100$.\newline
Plan Implementation: [\newline
  \{\newline
    "step\_number": "1",\newline
    "step": "Carefully read and analyze the given problem statement to understand the key elements involved, such as the binomial distribution and the specified condition $|Y/n - 0.25| < 0.05$.",\newline
    "reasoner": "LLM",\newline
    "output": "$<$step$><$number$>$1$<$/number$><$output$>$We are given that the random variable $Y$ follows a binomial distribution $b(n, 0.25)$, where $n$ is the number of trials and 0.25 is the probability of success in each trial. The problem requires us to find a lower bound for the probability $P(|Y/n - 0.25| < 0.05)$ when $n = 100$. This condition can be rewritten as $0.2 < Y/n < 0.3$, which is equivalent to $20 < Y < 30$. Therefore, we need to calculate $P(20 < Y < 30)$ for $Y \sim \text{Binomial}(100, 0.25)$.$<$/output$><$/step$>$",\newline
    "error": null\newline
  \},\newline
  \{\newline
    "step\_number": "2",\newline
    "step": "Identify whether Chebyshev's inequality should be applied, considering the need to find a lower bound for the probability $P(-5 < Y - 25 < 5)$.",\newline
    "reasoner": "LLM",\newline
    "output": "$<$step$><$number$>$2$<$/number$><$output$>$Chebyshev's inequality is suitable for finding a lower bound on probabilities related to the deviation of a random variable from its mean. Given the need to determine a lower bound for $P(-5 < Y - 25 < 5)$, where $Y = 25$ is the expected value of the binomial distribution $Y \sim b(100, 0.25)$, Chebyshev's inequality can be applied. The inequality states that for any $k > 0$, $P(|Y - 25| < 5) \geq 1 - \frac{\sigma^2}{k^2}$, where $\sigma^2$ is the variance of the distribution.$<$/output$><$/step$>$",\newline
    "error": null\newline
  \},\newline
  \{\newline
    "step\_number": "3",\newline
    "step": "Calculate the mean $\mu_Y$ and variance $\sigma_Y^2$ of the binomial distribution $Y$ to gather necessary parameters.",\newline
    "reasoner": "LLM",\newline
    "output": "$<$step$><$number$>$3$<$/number$><$output$>$The mean $\mu_Y$ and variance $\sigma_Y^2$ of a binomial distribution $Y \sim b(n, p)$ are calculated as follows:\newline- Mean $\mu_Y = n \cdot p = 100 \cdot 0.25 = 25$.\newline- Variance $\sigma_Y^2 = n \cdot p \cdot (1 - p) = 100 \cdot 0.25 \cdot 0.75 = 18.75$.$<$/output$><$/step$>$",\newline
    "error": null\newline
  \},\newline
  \{\newline
    "step\_number": "4",\newline
    "step": "Translate the condition $|Y/n - 0.25| < 0.05$ using variance-based analysis to align with the problem's requirements.",\newline
    "reasoner": "LLM",\newline
    "output": "$<$step$><$number$>$4$<$/number$><$output$>$The condition $|Y/n - 0.25| < 0.05$ translates to $0.2 < Y/n < 0.3$. Multiplying through by $n = 100$, we have $20 < Y < 30$, which aligns with the requirement to apply Chebyshev's inequality. In terms of deviation from the mean, this is equivalent to $P(-5 < Y - \mu_Y < 5)$, where $\mu_Y = 25$ is the mean we calculated earlier. This allows us to apply variance-based methods like Chebyshev's inequality to find a probability bound.$<$/output$><$/step$>$",\newline
    "error": null\newline
  \},\newline
  \end{tcolorbox}
  \begin{tcolorbox}[colback=gray!10, colframe=black, title=]
\footnotesize
  \{\newline
    "step\_number": "5",\newline
    "step": "Calculate the standard deviation $\sigma_Y$ from the variance $\sigma_Y^2$.",\newline
    "reasoner": "LLM",\newline
    "output": "$<$step$><$number$>$5$<$/number$><$output$>$The standard deviation $\sigma_Y$ is the square root of the variance $\sigma_Y^2$. Therefore, $\sigma_Y = \sqrt{18.75} \approx 4.33$.$<$/output$><$/step$>$",\newline
    "error": null\newline
  \},\newline
  \{\newline
    "step\_number": "6",\newline
    "step": "Determine the value of $k$ using the relation $k\sigma_Y = 5$ as part of the variance-based form.",\newline
    "reasoner": "LLM",\newline
    "output": "$<$step$><$number$>$6$<$/number$><$output$>$The value of $k$ is determined by the equation $k \cdot \sigma_Y = 5$, where $\sigma_Y \approx 4.33$. Thus, $k = \frac{5}{4.33} \approx 1.1557$.$<$/output$><$/step$>$",\newline
    "error": null\newline
  \},\newline
  \{\newline
    "step\_number": "7",\newline
    "step": "Apply Chebyshev's inequality using the parameters calculated to find the lower bound for the specified probability.",\newline
    "reasoner": "Tools: Python Executor",\newline
    "output": "0.25",\newline
    "error": null\newline
  \},\newline
  \{\newline
    "step\_number": "8",\newline
    "step": "Compute the numerical lower bound to obtain the final answer, ensuring precision in calculations.",\newline
    "reasoner": "Tools: Python Executor",\newline
    "output": "0.25",\newline
    "error": null\newline
  \},\newline
  \{\newline
    "step\_number": "9",\newline
    "step": "Revisit the problem statement to ensure all aspects are addressed with the results.",\newline
    "reasoner": "LLM",\newline
    "output": "$<$step$><$number$>$9$<$/number$><$output$>$The problem asked for a lower bound for $P(|Y/n - 0.25| < 0.05)$ with $Y \sim b(100, 0.25)$. We translated this to $P(20 < Y < 30)$, equivalent to $P(-5 < Y - 25 < 5)$ Using Chebyshev's inequality, we calculated that this probability is at least 0.25. All aspects of the problem statement have been addressed and calculated accurately, ensuring the lower bound provided is precise.$<$/output$><$/step$>$",\newline
    "error": null\newline
  \},\newline
  \{\newline
    "step\_number": "10",\newline
    "step": "Format and output the final answer according to specified output requirements.",\newline
    "reasoner": "Tools: Python Executor",\newline
    "output": "0.25",\newline
    "error": null\newline
  \},\newline
  \{\newline
    "step\_number": "11",\newline
    "step": "FIN",\newline
    "output": "FINISHED",\newline
    "error": null\newline
  \}\newline
]
\end{tcolorbox}

\subsection{Error Analysis}
In this section, we analyze the performance of RunAgent plan implementation and GPT-4o plan implementation on the Natural Plan Calendar Scheduling Dataset using bootstrapping and the McNemar statistical test. 

\begin{figure}
    \centering
    \includegraphics[width=\linewidth]{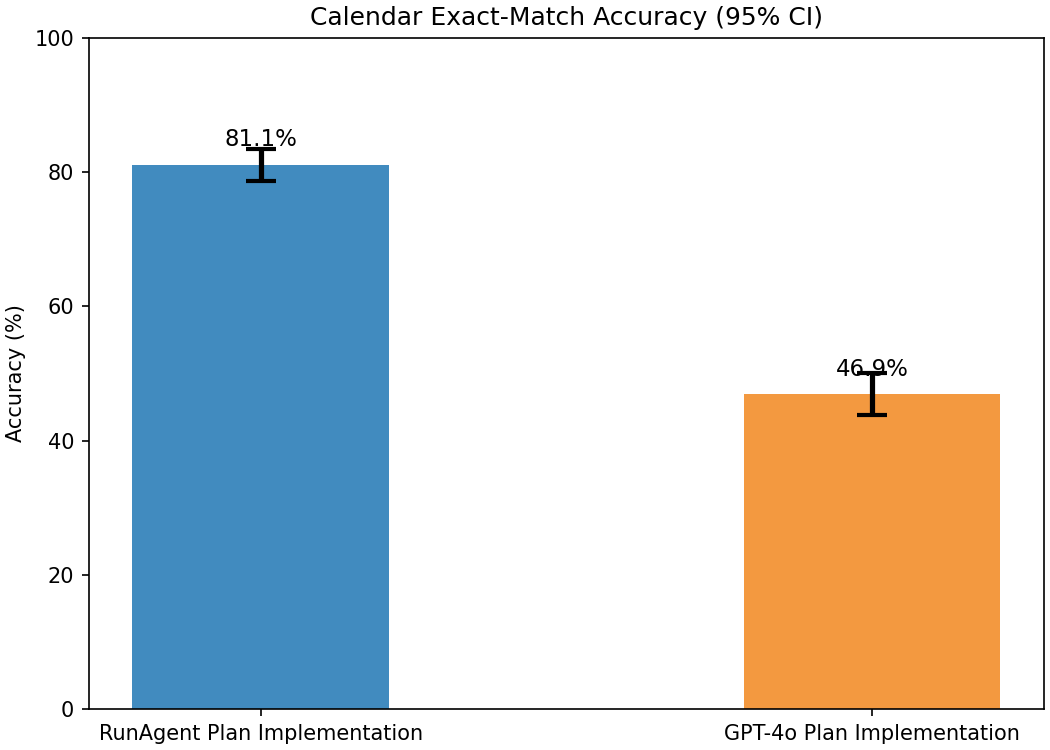}
    \caption{$95\%$ Confidence intervals for Natural Plan Calendar Scheduling.}
    \label{fig: calendar scheduling confidence intervals}
\end{figure}

In Fig.~\ref{fig: calendar scheduling confidence intervals}, we plot the $95\%$ confidence intervals for both plan implementation methods after using bootstrapping. We observe that RunAgent performs significantly better than GPT-4o when implementing the same plan.

In Table~\ref{tab: McNemar results}, we have the contingency table for the McNemar test. The null hypothesis in this case is that both methods are expected to exhibit the same performance on a randomly chosen instance. The alternative hypothesis is that RunAgent is systematically better than GPT-4o in plan implementation. Using this test, we can calculate that the chi-square value is $253.17$ and the corresponding p-value is $5.29 \times 10^{-56}$, demonstrating that RunAgent performs better than GPT-4o for plan implementation with very high certainty.

\begin{table}[t]
    \centering
    \small
    \begin{tabular}{|c|c|c|c|}
    \hline
      & GPT-4o Corr. & GPT-4o Incorr. & Total\\
    \hline
    RunAgent Corr. & 409 & 402 & 811\\
    \hline
    RunAgent Incorr. & 60 & 129 & 189\\
    \hline
    Total & 469 & 531 & 1000\\
    \hline
    \end{tabular}
    \caption{\textbf{McNemar Test Contingency Table}}
    \label{tab: McNemar results}
\end{table}

\subsection{Examples of Facts}

\subsubsection{Example for Natural Plan Calendar Scheduling}
\begin{itemize}
    \item[a)] If a busy slot ends at time t, then the person is considered free at time t. For example, if the person is busy from 10:00 AM to 12:00 PM, then the person is considered free at 12:00 PM, and a meeting can be scheduled at 12:00 PM for this person.
    \item[b)] There is no need to have breaks between work for any person.
    \item[c)] A solution always exists.
    \item[d)] Each time interval is considered as half open with the open part being the end of the interval.
\end{itemize}

\subsubsection{Example for Natural Plan Trip Planning}
\begin{enumerate}
    \item[a)] If you spend any time of the day in a city, count it as a full day spent in that city.
    \item[b)] If you spend a day in two cities, count the day for both cities.
    \item[c)] The day you arrive in a city is counted as a full day spent in the city, and the trip starts on the day you arrive.
    \item[d)] The day you leave a city is counted as a full day spent in the city, and the trip ends on the day you leave.
    \item[e)] If you spend days x to y in a city, then you spent y-x+1 days in that city. So if you spend days 3-6 in the city, you spent 4 days in that city.
    \item[f)] First find the order in which you visit the cities in the trip, then find the days you spend in each city.
\end{enumerate}

\subsubsection{Example for Scibench Stat}
\begin{enumerate}
    \item[a)] Chebyshev's inequality should be used when applicable.
\end{enumerate}

\end{document}